\documentclass{article}
\usepackage{spconf,amsmath,graphicx}
\usepackage{color}
\usepackage{xcolor}
\usepackage{amssymb}
\usepackage{multicol}
\usepackage{multirow}
\definecolor{cvprblue}{rgb}{0.21,0.49,0.74}
\usepackage[colorlinks, urlcolor=black, citecolor=cvprblue]{hyperref}

\usepackage{enumitem}
\setlist{nosep, leftmargin=14pt}

\usepackage{mwe} 

\def\etal{{\em et al.~}}

\title{Semi- and Weakly-Supervised Learning for Mammogram Mass Segmentation with Limited Annotations}
\name{Xinyu Xiong$^{1\star}$\thanks{$^\star$ Work done when interning at Deepwise AI Lab.}, Churan Wang$^{2\dagger}$, Wenxue Li$^3$, Guanbin Li$^{1 , 4\dagger}$\thanks{$^\dagger$ Corresponding Authors.}}
\address{$^1$School of Computer Science and Engineering, Sun Yat-sen University, Guangzhou, China\\
$^2$School of Computer Science, Peking University, Beijing, China \\
$^3$School of Future Technology, Tianjin University, Tianjin, China\\
$^4$Research Institute, Sun Yat-sen University, Shenzhen, China
}

\begin{document}
\maketitle
\begin{abstract}
Accurate identification of breast masses is crucial in diagnosing breast cancer; however, it can be challenging due to their small size and being camouflaged in surrounding normal glands. Worse still, it is also expensive in clinical practice to obtain adequate pixel-wise annotations for training deep neural networks. To overcome these two difficulties with one stone, we propose a semi- and weakly-supervised learning framework for mass segmentation that utilizes limited strongly-labeled samples and sufficient weakly-labeled samples to achieve satisfactory performance. The framework consists of an auxiliary branch to exclude lesion-irrelevant background areas, a segmentation branch for final prediction, and a spatial prompting module to integrate the complementary information of the two branches. We further disentangle encoded obscure features into lesion-related and others to boost performance. Experiments on CBIS-DDSM and INbreast datasets demonstrate the effectiveness of our method.
\end{abstract}


\section{Introduction}

With approximately 2.3 million new cases in 2020~\cite{GLOBOCAN}, female breast cancer has replaced lung cancer as the most commonly diagnosed cancer, posing a severe threat to human health~\cite{wang2021bilateral,wang2020br}. 
Screening mammograms for mass analysis play a crucial role in the early detection of breast cancer and can effectively reduce mortality~\cite{wang2021dae,wang2022disentangling,wang2022learning}.
However, this process can be tedious and error-prone~\cite{BJC13_BENEFIT}, leading to increased demand for computer-aided diagnosis systems. 
Recent mammogram mass segmentation methods~\cite{PMB20_AUNet,CBM21_FSUNet,MLMI23_MammoSAM} have shown promising progress thanks to the rapid development of deep learning. 
Nevertheless, these models are data-hungry and rely heavily on high-quality annotations that are costly to obtain in clinical practice. Therefore, label-efficient learning methods are necessary to advance toward real-world applications.

\begin{figure}[t]
	\centering
    \includegraphics[width=0.48\textwidth]{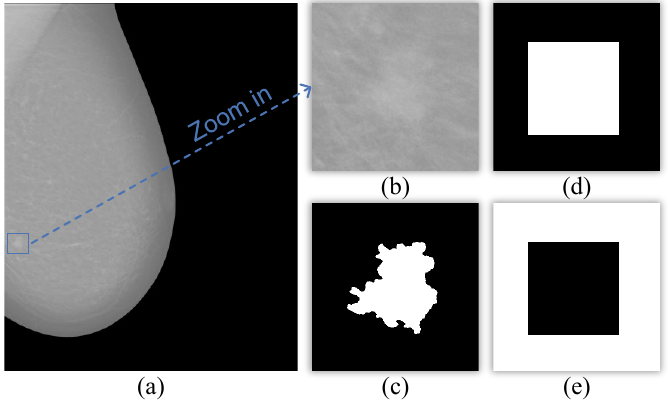}
	\caption{(a) A whole mammogram with mass location marked in blue. (b) Magnified view of the mass. (c) Magnified view of the corresponding strong label. (d) Magnified view of the corresponding weak label. (e) Magnified view of the reversed weak label. In weak labels, we consider the foreground as uncertain areas and the background as definite normal glands.} 
	\label{fig:title}
\end{figure}

Semi- and weakly-supervised learning~\cite{PR22_semiweakly}, also known as hybrid-supervised learning~\cite{AAAI22_DCR,SPL23_HybridVPS}, aims to train models with limited strong labels (e.g., pixel-wise masks) and sufficient cheap labels (e.g., bounding boxes.) This learning paradigm combines semi-~\cite{ICCV23_ESL} and weak-supervision advantages, enabling label-efficient learning. Since box annotation does not contain accurate boundary information that can distinguish lesions from the background, a critical issue in hybrid supervision is how to eliminate noise associated with weak labels during learning. To address this, a dual-decoder framework was proposed in~\cite{ECCV20_DualBranch} to reduce mutual interference between strong and weak labels. Pan~\etal\cite{AAAI22_DCR} developed a dynamic instance indicator to re-weight samples of different quality, while self-correcting modules were developed in~\cite{CVPR20_SelfCorr} to improve the accuracy of generated pseudo labels.

\begin{figure*}[t]
	\centering
	\includegraphics[width=0.9\textwidth]{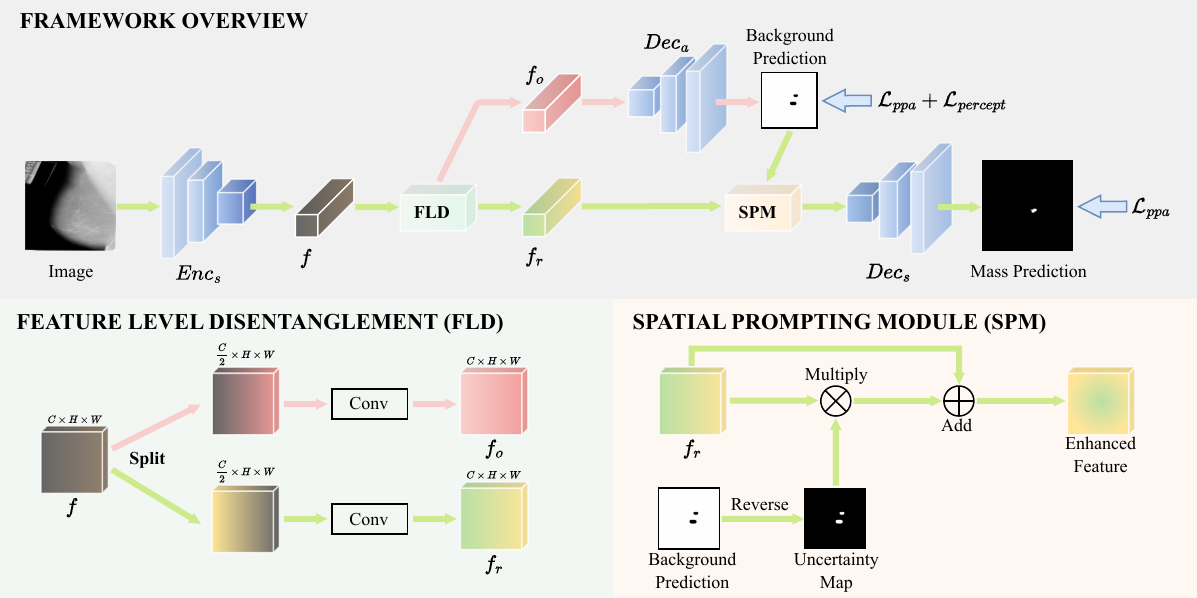}
	\caption{Overview of the hybrid-supervised segmentation framework. The green line indicates the shared data flow, and the pink line indicates the data flow of strong set. The background decoder $Dec_{a}$ is directly supervised by reversed bounding box masks $y^w$, and the segmentation decoder $Dec_s$ is supervised by fine labels $y^s$.} 
	\label{fig:overview}
\end{figure*}

While previous methods have made considerable progress, many of them treat predictions from weak labels as pseudo-labels and expect them to be consistent with strong labels after regularization, operating in a ``consistency" manner. 
However, such a strategy shall face challenges on mammograms because the boundaries between masses and normal glands are often unclear, as shown in Fig.~\ref{fig:title}, making it difficult to regularize. This provoked a thought, ``can we solve the two problems of annotation shortage and accurate segmentation in a unified framework?"

To break through the difficulties, we draw inspiration from the two-stage process for radiologists locating and segmentation the mass. He or she will first find several suspicious areas from a whole mammogram. These areas are then further examined to determine whether a lesion and corresponding boundary are present. From here, it can find that these are two different tasks: 1) The first stage is a classification task, dividing the whole image into the definite background and the uncertain area. 2) The second stage is a regression task, focusing on accurately refining the confidence value of pixels in the uncertain area to identify lesion boundaries better. Motivated by the above analysis, we propose a label-efficient mammogram mass segmentation framework, the contribution of which is two-fold: 

\begin{itemize}
    \item \textbf{Task Perspective}. We provide a novel perspective that mammogram mass segmentation could be disentangled into two tasks, and such disentanglement is helpful for performance improvement.
    
    \item \textbf{Feature Perspective}. We further disentangle the obscure features extracted by the encoder into lesion-related features and others. A spatial prompting module is designed to enhanced lesion-related features, contributing to identify mass boundaries in uncertain areas more accurately.
\end{itemize}

\section{Method}
The problem of hybrid-supervised learning is formalized as follows. The entire training dataset $\mathcal{D}$ consists of a strong subset $\mathcal{D}_{S} = \{x_i^s, y_i^s\}_{i=1}^{M}$ with $M$ samples  and a weak subset $\mathcal{D}_{W} = \{x_i^w, y_i^w\}_{i=1}^{N}$ with $N$ samples, where $x$, $y^s$ and $y^w$ denote input images, pixel-level fine labels and reversed bounding box labels (Fig.~\ref{fig:title}(e)), respectively. Our goal is to train an effective segmentation model by seamlessly integrating strong and weak labels with different levels of semantics.

\subsection{Task Level Disentanglement} 
As shown in Fig.~\ref{fig:overview}, our hybrid-supervised learning framework contains a shared feature encoder $Enc_s$, a segmentation decoder $Dec_s$, and an auxiliary background decoder $Dec_{a}$. The encoder adopts the pre-trained ResNet-34~\cite{CVPR16_ResNet}, and both decoders follow the architecture of U-Net decoder~\cite{MICCAI15_UNet}.

Inspired by the two stages of the manual diagnosis process, our method works in a multi-task manner. Specifically, these two decoders differ in the supervision signals they receive. The background decoder $Dec_{a}$ receives direct supervision through reversed bounding box masks $y^w$, while the segmentation decoder $Dec_s$ is trained using pixel-wise masks $y^s$. Such task-level disentanglement effectively aligns the characteristics of different labels with the specific aims of two decoders. As $y^w$ only capture the definitive background area, they serve to facilitate $Dec_{a}$ in distinguishing between the definitive background and uncertain areas that potentially contain lesions. Conversely, given their awareness of the definite foreground area, $y^s$ are employed to supervise $Dec_s$ and guide it toward precise identification of lesion boundaries.

\subsection{Feature Level Disentanglement}

Owing to two inherent characteristics of breast mass segmentation, the encoded raw features $f$ lack distinctiveness between masses and normal glands. Firstly, masses typically only occupy a small fraction of the entire image. Secondly, the boundaries between the masses and the background are often indistinct. Consequently, directly using identical features $f$ with obscure semantics across two tasks (mass segmentation and background identification) results in feature interference and diminished performance.

To tackle this challenge, we propose to disentangle encoded features $f$ into distinct sets: lesion-related features $f_r$ and others $f_o$. Initially, $f$ are bifurcated along the channel dimension. Subsequently, the divided features pass separate $1 \times 1$ convolution layers to upsampling back to their original dimensions. These disentangled features are then directed into respective decoders $Dec_s$ and $Dec_a$, effectively mitigating feature interference. This design enables each decoder to concentrate on its specific task without being impeded by obscure semantics.

\subsection{Spatial Prompting Module}

To harness the potential lesion location cues offered by the background branch $Dec_a$, we introduce a spatial prompting module, the core insight of which is emulating the human specialist in delineating the precise boundaries of lesions within uncertain areas.

Specifically, since negative areas in the background predictions correspond to the uncertain areas that contain suspected lesions, we reverse the predictions and obtain uncertainty maps. The uncertainty maps are multiplied with lesion-related features $f_r$, prompting the segmentation decoder $Dec_s$ to focus on intricate lesion details within uncertain areas.

\subsection{Loss Function}
Since breast masses are generally small in size, we adapt the pixel-position aware loss~\cite{AAAI20_F3Net}, as shown in Eq.~\ref{eq:ppa}, to supervise the segmentation branch: \begin{equation}
\label{eq:ppa}
\mathcal{L}_{ppa} = \mathcal{L}^{w}_{bce} + \mathcal{L}^{w}_{iou}.
\end{equation} 

For the auxiliary branch, an intuitive approach is to use the same loss function as the segmentation branch~\cite{ECCV20_DualBranch}. However, it can be found that the objectives of these two branches are different. Specifically, for the auxiliary branch, its uncertain area recall rate performance is more critical than other metrics. Since the role of the auxiliary branch is to guide the segmentation branch to exclude as much as possible irrelevant background area, it is acceptable to misidentify some of the background pixels as uncertain pixels but not vice versa, as this would mislead the segmentation branch.

To encourage the auxiliary branch to preserve as many potential mass pixels as possible, we introduce an uncertain area perception loss, which can be defined as: \begin{equation}
\label{eq:recall}
\mathcal{L}_{percept} = 1 - \frac{TN}{TN + FP},
\end{equation} where FP and TN denote false positives and true negatives, respectively. It is important to note that as the auxiliary branch is supervised by reversed weak labels, therefore positive predictions represent background areas and negative predictions represent uncertain areas. Involving $\mathcal{L}_{percept}$ in supervising the auxiliary background branch helps improve its ability to perceive potential mass areas.

Finally, our total loss function is defined as follows: \begin{equation}
\label{eq:fin}
\mathcal{L}_{total} = \mathcal{L}_{ppa}^{seg} + \mathcal{L}_{ppa}^{aux} + \mathcal{L}_{percept}^{aux},
\end{equation} where $\mathcal{L}_{ppa}^{seg}$, $\mathcal{L}_{ppa}^{aux}$, and $\mathcal{L}_{percept}^{aux}$ denote pixel-position aware loss for the segmentation branch, auxiliary branch, and uncertain area perception loss for the auxiliary branch, respectively.

\section{Experiment}
\subsection{Datasets} 
We conduct experiments on the two following publicly available mammography datasets:

\textbf{CBIS-DDSM}~\cite{CBIS} dataset contains 1592 mass images with both pixel-wise masks and bounding boxes. We randomly divided it into 80\%:20\% for training and testing. For the training set, 10\% samples are strong-supervised, and the remaining are weak-supervised unless otherwise stated.

\textbf{INbreast}~\cite{INBREAST} dataset contains 107 mass images in total. It serve as a test set to evaluate cross-domain generalizability in our experiments.

\subsection{Implementation Details}
Our model is implemented with Pytorch and trained on a single NVIDIA RTX 2080Ti. We employ the Adam optimizer with an initial learning rate of 0.0001, and a poly learning rate decay policy is applied to stabilize training, which is $lr=lr_{init}\times(1-\frac{epoch}{nEpoch})^{power}$, where $power=0.9$, $nEpoch=50$.
The batch size is set to 8. All the images are resized to 352$\times$352. Two data augmentation strategies are employed, including random vertical and horizontal flips.

\subsection{Compared Methods}
Our approach is compared with the following state-of-the-art hybrid-supervised methods, including Strong-Weak~\cite{ECCV20_DualBranch}, Self-Correcting~\cite{CVPR20_SelfCorr}, Marco-Micro~\cite{MICCAI20_MAMI}, and DCR~\cite{AAAI22_DCR}. In addition, three baselines are also involved in the discussion, including Strong-Only (training with strongly labeled samples only), Weak-Only (training with weakly labeled samples only), and Vanilla-Hybrid (directly training with a mix of strong and weak labels). For quantitative comparison, we adopt Dice Coefficient and Structure Measure ($S_m$)~\cite{ICCV17_Sm}. 

\begin{table}[t]
\centering
\caption{Quantitative comparison on the CBIS-DDSM~\cite{CBIS} and INbreast~\cite{INBREAST} datasets. Results are shown in percentages. The best \textit{hybrid-supervised} results are highlighted in \textbf{bold}.}
\label{tab:comp}
\renewcommand\arraystretch{1.2}
\renewcommand\tabcolsep{2.0pt}
\begin{tabular}{c|c|c|c|c|c|c} 
\cline{1-7}
\multirow{2}{*}{Methods} & \multirow{2}{*}{S} & \multirow{2}{*}{W} & \multicolumn{2}{c|}{CBIS-DDSM} & \multicolumn{2}{c}{INbreast}  \\ 
\cline{4-7}
 & & & $Dice$ & $S_m$ & $Dice$ & $S_m$          \\ 
\hline
Strong-Only & 100\% & 0\% & 75.48 & 85.25 & 63.98 & 78.83 
  \\
Strong-Only & 10\% & 0\% & 65.66 & 80.05 & 47.64 & 70.31  \\
Weak-Only & 0\% & 100\% & 64.87 & 77.35 & 53.12 & 70.62
 \\
Vanilla-Hybrid & 10\% & 90\% & 65.03 & 77.54 & 53.46 & 70.46 \\
Strong-Weak & 10\% & 90\% & 69.31 & 81.60 & 54.21 & 73.74 \\
Self-Correcting & 10\% & 90\% & 67.23 & 80.68 & 54.16 & 73.54  \\
Marco-Micro & 10\% & 90\% & 67.89 & 79.87 & 53.80 & 72.07   \\
DCR & 10\% & 90\% & 68.74 & 81.65 &   57.66 & 74.88 \\ 
\hline
\textbf{Ours} & \textbf{10\%} & \textbf{ 90\%} & \textbf{72.26} & \textbf{83.18} & \textbf{60.11} & \textbf{76.02} \\
\hline
\end{tabular}
\end{table}

\subsection{Result Analysis} 

\textbf{Intra-Domain Comparison} results are shown in the ``CBIS-DDSM" row of Table~\ref{tab:comp}. It can be found that Vanilla-Hybrid obtains lower results than Stong-Only, demonstrating that direct treat weak labels as strong ones will introduce much noise, which leads to performance degradation. Therefore, it is necessary to design hybrid supervision methods to handle weak labels correctly. Among the different strategies, our method achieves state-of-the-art results on all metrics, demonstrating the superiority of our framework. 

\textbf{Cross-Domain Validation} results are shown in the ``INbreast" row of Table~\ref{tab:comp}. To assess the generalization capability, we utilize the previously trained models on the CBIS-DDSM dataset for evaluation on the INbreast dataset. Our method still performs best with the Dice score of 60.11\%.

\begin{table}[t]
\centering
\caption{Ablation study on the CBIS-DDSM~\cite{CBIS} and INbreast~\cite{INBREAST} datasets. FD, SPM, and PL are abbreviations of Feature Level Disentanglement, Spatial Prompting Module, and Perception Loss, respectively.}
\label{tab:abla}
\renewcommand\arraystretch{1.2}
\renewcommand\tabcolsep{2.0pt}
\begin{tabular}{c|c|c|c|c|c|c} 
\cline{1-7}
\multirow{2}{*}{Methods} & \multirow{2}{*}{S} & \multirow{2}{*}{W} & \multicolumn{2}{c|}{CBIS-DDSM} & \multicolumn{2}{c}{INbreast}  \\ 
\cline{4-7}
 & & & $Dice$ & $S_m$ & $Dice$ & $S_m$           \\ 
\hline
Ours & 7.5\% & 92.5\% & 67.74 & 74.12   
 & 54.85 & 61.82    
 \\
Ours & 10\% & 90\% & 72.26 & 83.18 &  60.11 & 76.02    
\\
Ours & 20\% & 80\% & 73.67 & 84.04  & 60.34 & 76.65 
\\
Ours & 50\% & 50\% & 75.26 & 84.97  & 62.08 & 77.41  
\\
Ours (w/o FD) & 10\% & 90\% & 69.85 & 81.38 & 56.76 & 71.01    
\\
Ours (w/o SPM) & 10\% & 90\% & 68.43 & 81.45 & 54.80 & 72.87    
\\
Ours (w/o PL) & 10\% & 90\% & 68.35 & 80.93 &  57.73 & 73.59 
\\
\hline
\end{tabular}
\end{table}

\begin{figure}[!t]
	\centering
	\includegraphics[width=0.48\textwidth]{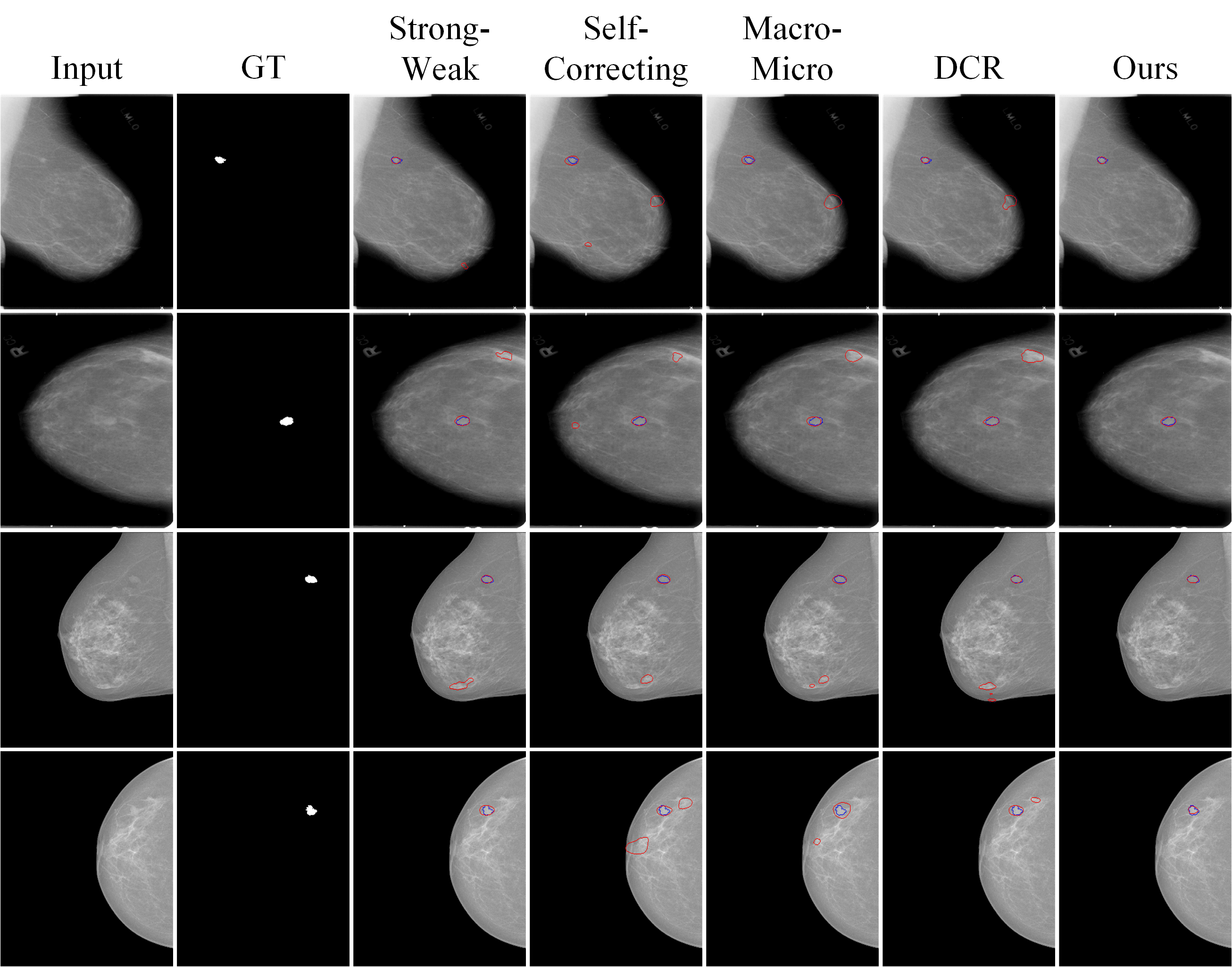}
	\caption{Visual comparison on mammogram
    mass segmentation. Ground Truths are marked in blue and prediction results are marked in red. Zoom in for best view.} 
	\label{fig:visual}
\end{figure}

\textbf{Visual Comparison} results are shown in Fig.~\ref{fig:visual}. Our method can obtain more accurate segmentation results in various cases compared with other methods.

\subsection{Ablation Study}
We conduct ablation experiments as presented in Table.~\ref{tab:abla}. 

\textbf{Performance Under Different Strong Labels.} It can be found that more strong labels generally lead to higher results. We choose to use 10\% strong labels because it can better balance performance and annotation budget.

\textbf{Effectiveness of Different Strategies.} Removing the feature level disentanglement mechanism, spatial prompting module, perception loss resulting in a 2.41\%, 3.83\%, 3.91\% decrease in the Dice score on the CBIS-DDSM dataset, demonstrating the effectiveness of our proposed strategies.

\section{Conclusion}
This paper presents a hybrid-supervised framework for breast mass segmentation, which can alleviate tedious annotation efforts by utilizing sufficient weak bounding box labels. Inspired by the manual inspection process, we disentangle the segmentation task, where weakly labeled samples serve as perceiving potential locations of lesions, and strongly labeled ones focus on identifying the accurate boundary of lesions. 
Experiments demonstrate the superiority of our framework.

\section{Compliance with Ethical Standards} This research study was conducted retrospectively using human subject data made available in open access by~\cite{CBIS,INBREAST}. Ethical approval was not required as confirmed by the license attached with the open access data.

\section{Acknowledgments}
This work was supported in part by the National Natural Science Foundation of China (NO.~62322608), in part by the Open Project Program of the Key Laboratory of Artificial Intelligence for Perception and Understanding, Liaoning Province (AIPU, No.~20230003), and in part by the Shenzhen Science and Technology Program (NO.~JCYJ2022053014121\\1024). 

{
\small
\bibliographystyle{IEEEbib}
\bibliography{refs}

\begin{thebibliography}{10}

\bibitem{GLOBOCAN}
Hyuna Sung, Jacques Ferlay, Rebecca~L Siegel, Mathieu Laversanne, Isabelle Soerjomataram, Ahmedin Jemal, and Freddie Bray,
\newblock ``Global cancer statistics 2020: Globocan estimates of incidence and mortality worldwide for 36 cancers in 185 countries,''
\newblock {\em CA: a cancer journal for clinicians}, vol. 71, no. 3, pp. 209--249, 2021.

\bibitem{wang2021bilateral}
Churan Wang, Jing Li, Fandong Zhang, Xinwei Sun, Hao Dong, Yizhou Yu, and Yizhou Wang,
\newblock ``Bilateral asymmetry guided counterfactual generating network for mammogram classification,''
\newblock {\em IEEE Transactions on Image Processing}, vol. 30, pp. 7980--7994, 2021.

\bibitem{wang2020br}
Chu-ran Wang, Fandong Zhang, Yizhou Yu, and Yizhou Wang,
\newblock ``Br-gan: Bilateral residual generating adversarial network for mammogram classification,''
\newblock in {\em MICCAI}. Springer, 2020, pp. 657--666.

\bibitem{wang2021dae}
Churan Wang, Xinwei Sun, Fandong Zhang, Yizhou Yu, and Yizhou Wang,
\newblock ``Dae-gcn: Identifying disease-related features for disease prediction,''
\newblock in {\em MICCAI}. Springer, 2021, pp. 43--52.

\bibitem{wang2022disentangling}
Chu-ran Wang, Fei Gao, Fandong Zhang, Fangwei Zhong, Yizhou Yu, and Yizhou Wang,
\newblock ``Disentangling disease-related representation from obscure for disease prediction,''
\newblock in {\em ICML}. PMLR, 2022, pp. 22652--22664.

\bibitem{wang2022learning}
Churan Wang, Jing Li, Xinwei Sun, Fandong Zhang, Yizhou Yu, and Yizhou Wang,
\newblock ``Learning domain-agnostic representation for disease diagnosis,''
\newblock in {\em ICLR}, 2022.

\bibitem{BJC13_BENEFIT}
Michael~G Marmot, DG~Altman, DA~Cameron, JA~Dewar, SG~Thompson, and Maggie Wilcox,
\newblock ``The benefits and harms of breast cancer screening: an independent review,''
\newblock {\em British journal of cancer}, vol. 108, no. 11, pp. 2205--2240, 2013.

\bibitem{PMB20_AUNet}
Hui Sun, Cheng Li, Boqiang Liu, Zaiyi Liu, Meiyun Wang, Hairong Zheng, David~Dagan Feng, and Shanshan Wang,
\newblock ``Aunet: Attention-guided dense-upsampling networks for breast mass segmentation in whole mammograms,''
\newblock {\em Physics in Medicine \& Biology}, vol. 65, no. 5, pp. 055005, 2020.

\bibitem{CBM21_FSUNet}
Jiande Pi, Yunliang Qi, Meng Lou, Xiaorong Li, Yiming Wang, Chunbo Xu, and Yide Ma,
\newblock ``Fs-unet: Mass segmentation in mammograms using an encoder-decoder architecture with feature strengthening,''
\newblock {\em Computers in Biology and Medicine}, vol. 137, pp. 104800, 2021.

\bibitem{MLMI23_MammoSAM}
Xinyu Xiong, Churan Wang, Wenxue Li, and Guanbin Li,
\newblock ``Mammo-sam: Adapting foundation segment anything model for automatic breast mass segmentation in whole mammograms,''
\newblock in {\em MLMI}. Springer, 2023, pp. 176--185.

\bibitem{PR22_semiweakly}
Yude Wang, Jie Zhang, Meina Kan, and Shiguang Shan,
\newblock ``Learning pseudo labels for semi-and-weakly supervised semantic segmentation,''
\newblock {\em Pattern Recognition}, vol. 132, pp. 108925, 2022.

\bibitem{AAAI22_DCR}
Junwen Pan, Qi~Bi, Yanzhan Yang, Pengfei Zhu, and Cheng Bian,
\newblock ``Label-efficient hybrid-supervised learning for medical image segmentation,''
\newblock in {\em AAAI}, 2022, vol.~36, pp. 2026--2034.

\bibitem{SPL23_HybridVPS}
Wenxue Li, Xinyu Xiong, Siying Li, and Fugui Fan,
\newblock ``Hybridvps: Hybrid-supervised video polyp segmentation under low-cost labels,''
\newblock {\em IEEE Signal Processing Letters}, 2023.

\bibitem{ICCV23_ESL}
Jie Ma, Chuan Wang, Yang Liu, Liang Lin, and Guanbin Li,
\newblock ``Enhanced soft label for semi-supervised semantic segmentation,''
\newblock in {\em ICCV}, 2023, pp. 1185--1195.

\bibitem{ECCV20_DualBranch}
Wenfeng Luo and Meng Yang,
\newblock ``Semi-supervised semantic segmentation via strong-weak dual-branch network,''
\newblock in {\em ECCV}. Springer, 2020, pp. 784--800.

\bibitem{CVPR20_SelfCorr}
Mostafa~S Ibrahim, Arash Vahdat, Mani Ranjbar, and William~G Macready,
\newblock ``Semi-supervised semantic image segmentation with self-correcting networks,''
\newblock in {\em CVPR}, 2020, pp. 12715--12725.

\bibitem{CVPR16_ResNet}
Kaiming He, Xiangyu Zhang, Shaoqing Ren, and Jian Sun,
\newblock ``Deep residual learning for image recognition,''
\newblock in {\em CVPR}, 2016, pp. 770--778.

\bibitem{MICCAI15_UNet}
Olaf Ronneberger, Philipp Fischer, and Thomas Brox,
\newblock ``U-net: Convolutional networks for biomedical image segmentation,''
\newblock in {\em MICCAI}. Springer, 2015, pp. 234--241.

\bibitem{AAAI20_F3Net}
Jun Wei, Shuhui Wang, and Qingming Huang,
\newblock ``F$^3$net: Fusion, feedback and focus for salient object detection,''
\newblock in {\em AAAI}, 2020, vol.~34, pp. 12321--12328.

\bibitem{CBIS}
Rebecca~Sawyer Lee, Francisco Gimenez, Assaf Hoogi, Kanae~Kawai Miyake, Mia Gorovoy, and Daniel~L Rubin,
\newblock ``A curated mammography data set for use in computer-aided detection and diagnosis research,''
\newblock {\em Scientific data}, vol. 4, no. 1, pp. 1--9, 2017.

\bibitem{INBREAST}
In{\^e}s~C Moreira, Igor Amaral, In{\^e}s Domingues, Ant{\'o}nio Cardoso, Maria~Joao Cardoso, and Jaime~S Cardoso,
\newblock ``Inbreast: toward a full-field digital mammographic database,''
\newblock {\em Academic radiology}, vol. 19, no. 2, pp. 236--248, 2012.

\bibitem{MICCAI20_MAMI}
Munan Ning, Cheng Bian, Donghuan Lu, Hong-Yu Zhou, Shuang Yu, Chenglang Yuan, Yang Guo, Yaohua Wang, Kai Ma, and Yefeng Zheng,
\newblock ``A macro-micro weakly-supervised framework for as-oct tissue segmentation,''
\newblock in {\em MICCAI}. Springer, 2020, pp. 725--734.

\bibitem{ICCV17_Sm}
Deng-Ping Fan, Ming-Ming Cheng, Yun Liu, Tao Li, and Ali Borji,
\newblock ``Structure-measure: A new way to evaluate foreground maps,''
\newblock in {\em ICCV}, 2017, pp. 4548--4557.

\end{thebibliography}
}
\end{document}